# Effective Image Tampering Localization with Multi-Scale ConvNeXt Feature Fusion

*Haochen Zhu* [1,2], *Gang Cao*[1,2], *Mo Zhao* [1,2]

[1]State Key Laboratory of Media Convergence and Communication, Communication University of China, Beijing 100024, China

[2]School of Computer and Cyber Sciences, Communication University of China, 100024 Beijing, China

*Abstract*—With the widespread use of powerful image editing tools, image tampering becomes easy and realistic. Existing image forensic methods still face challenges of low generalization performance and robustness. In this letter, we propose an effective image tampering localization scheme based on ConvNeXt network and multi-scale feature fusion. Stacked ConvNeXt blocks are used as an encoder to capture hierarchical multi-scale features, which are then fused in decoder for locating tampered pixels accurately. Combined loss and effective data augmentation are adopted to further improve the model performance. Extensive experimental results show that localization performance of our proposed scheme outperforms other state-of-the-art ones. The source code will be available at https://github.com/ZhuHC98/ITL-SSN.

## I. INTRODUCTION

With the development of image processing techniques, forgery images become rather realistic and difficult to be distinguished from real ones. Malicious manipulations, such as splicing, object copy-move, and removal, typically change the semantic expression of images. Misuse of tampered images would cause great harm to public security and judicial forensics. It is significant to develop forensic techniques for identifying forgery images and locating the tampered regions therein. The image tampering localization techniques can help us discover more refined forensic information, such as the answers to how and where the malicious alteration occurs.

In recent years, a number of image tampering localization schemes have been proposed by leveraging on deep learning architecture. Region- or pixel-level tampering localization has been achieved to some extent. Huh et al. [1] use image EXIF metadata as a monitoring signal to learn self-consistency by convolution neural network (CNN). The Noiseprint [2] and ForSim [3] algorithms work by predicting whether two image patches come from the same source by measuring the distance of learned features.

Such forensic schemes rely on checking consistency of specific clues and might be greatly affected by postprocessing. Some other schemes treat the image tampering localization as a pixel classification task [4-14], which is solved by leveraging on an end-to-end deep learning framework. The specific clue features and their regional inconsistency are no longer pursued. The tampering localization network is trained by massive tampered image samples with pixel-level tampering annotations. The deep learning architectures developed in prior works include deep CNN [4-8], two stream region-CNN [9], long short-term memory (LSTM) network [10, 11], and dense fully convolutional network (DFCN) [12], etc. Besides, spatial regional and boundary features extracted by DeepLab network [13] and self-attention mechanism [5, 14] are also applied. However, accuracy and robustness of such tampering localization algorithms are still insufficient for real applications.

In order to attenuate the deficiency of existing works, here we propose an effective image tampering localization scheme based on ConvNeXt network and multi-scale feature fusion in an encoder-decoder framework. Note that the complex artifacts incurred by various manipulations are hard to be modeled as specific artificial features. To capture such generic and inherent tampering features, ConvNeXt [15] which is one of the state-of-the-art representation learning networks is used as backbone encoder network. The hierarchical multi-scale features are extracted by stacked ConvNeXt blocks with large convolutional kernel and no-pooling structure, which benefit to learn discriminative and attentive representations. Furthermore, the tampered regions in real-world images typically vary in size. To adapt such scale variation, in decoder a top-down pathway inspired by Upernet [16] is exploited to fuse the multi-scale features for tampering localization. Besides, combined loss is adopted to guide effective model training, and data augmentation is used to enhance the generality and robustness. Performance advantages of the proposed tampering localization scheme have been verified extensively.

The rest of this letter is organized as follows. The proposed image tampering localization scheme is described in Section II, followed by experimental results and discussions in Section III. We draw the conclusions in Section IV.

## II. PROPOSED SCHEME

In this section, the proposed image tampering localization scheme is presented in detail. The encoder-decoder localization network based on deep CNNs is first introduced and discussed. Then the applied data augmentation strategy and loss function are described. An overview of the scheme is shown in Fig. 1.

*A. Encoder-Decoder Localization Network*

Let an input color image be denoted by $I \in \mathbb{R}^{H \times W \times 3}$, where $H$, $W$ are the image height and width, respectively. $I$ is the first input to a convolutional layer of 4×4 kernel with stride 4 for yielding an appropriate feature embedding $X_0 \in \mathbb{R}^{H/4 \times W/4 \times C}$, where $C$=128 is the number of feature map channels. The backbone encoder network owns four modules stacked with 3, 3, 27, 3 ConvNeXt blocks [15], respectively. Output of the $i$-th encoder module $E_i$ can be formulated as

$$X_i = E_i(X_{i-1}) \tag{1}$$

where $X_i \in \mathbb{R}^{H/2^{i+1} \times W/2^{i+1} \times 2^{i-1}C}$, $i$=1, 2, 3, 4. Three downsampling blocks with a rate of 1/2 are integrated into $E_2 \sim E_4$, respectively.

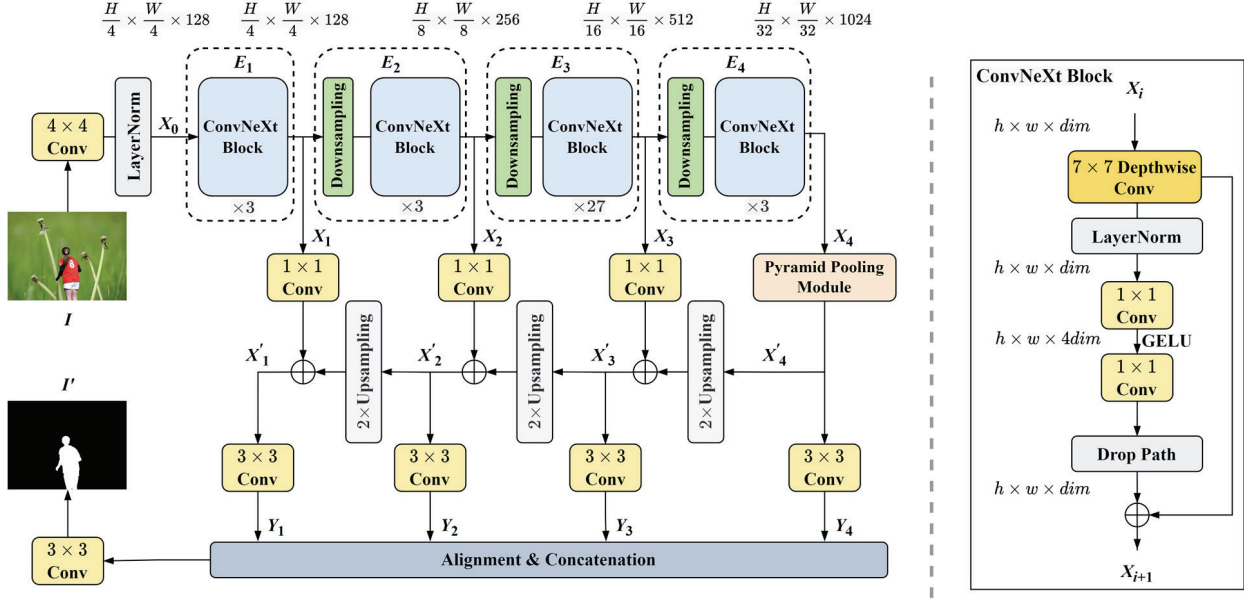

Fig. 1. Left: The proposed image tampering localization scheme. The encoder network consisting of ConvNeXt and downsampling blocks extracts features at different scales. The decoder network achieves tampering localization by fusing such multi-scale features. Right: Design for a ConvNeXt block.

As shown in the right of Fig. 1, a ConvNeXt block consists of 7×7 depthwise and 1×1 general convolutions. The former mainly mixes spatial domain information, and the latter extends and condenses the feature maps in channel dimension. The large convolutional kernel could provide a larger perceptual field for capturing large-scale and attentive tampering traces. Such complex large-kernel convolution operations are enforced with fewer channels for reducing model parameters. Efficient 1×1 convolutional layers are used to deepen the channels. As such, both global and local tampering trace features could be extracted cost-effectively by ConvNeXt encoder.

According to the hierarchical nature of the encoder network, low and high modules tend to capture local and large-scale complex patterns, respectively. The combined use of such low-level pixelwise and high-level semantic features would benefit to tampered region identification. To this end, multi-scale feature fusion in a top-down pathway is proposed in the decoder to estimate the pixel-level tampering localization map. The multi-scale features $\{X_i\}$ are adaptively fused via feature pyramid network (FPN) [17], which correspondingly yields a set of feature maps $Y_i$, $i$=4, 3, 2, 1. Specifically, $X_4$ is fed to a pyramid pooling module (PPM) [18], which empirically proposes an effective global prior representation, and converted into $Y_4$. As illustrated in Fig. 1, the high-level feature map $X'_{i+1}$ is upsampled by two times, denoted as $Up(X'_{i+1})$, and then merged with $X_i$ after 1×1 convolution operation $Conv_{1\times1}(\cdot)$. Such process is iterated and expressed as

$$X'_i = \begin{cases} Conv_{1\times1}(X_i) \oplus Up(X'_{i+1}), & i=1,2,3 \\ PPM[X_i], & i=4 \end{cases} \quad (2)$$

where $\oplus$ denotes the element-wise addition. Finally, the corresponding generated feature map $Y_i$ is

$$Y_i = Conv_{3\times3}(X'_i) \quad (3)$$

where $i$=1, 2, 3, 4. The upsampling and 1×1 convolution operations are used to align the two input arguments. A 3×3 convolution is followed by each merged map to eliminate aliasing effect.

Lastly, all scales of decoded feature maps $Y_i$, $i$=1, 2, 3, 4, are fused by performing bilinear upsampling alignment to $Y_1$ and concatenation. A 3×3 convolutional layer is subsequently enforced to further interface such fused feature and reduce its channel dimension. Finally, the tampering localization map $I'$ is yielded by predicting each pixel of $I$ as tampered or pristine.

*B. Data Augmentation*

Sufficient sample images are rather important for training the proposed tampering localization network. However, there lack of enough annotated real-world forgery images. Recent works [6, 10] have proposed to automatically synthesize plenty of forgery images based on object segments of COCO images [19]. In consideration of cost-effective computation, the object regions are randomly composited into pristine images. In order to simulate real-world post-processing, we propose to apply data augmentation operations to such composite sample images. Inspired by computer vision community, resizing, random cropping, and flipping are applied as geometric transformation. Then Gaussian noising, blurring, and photometric distortions of brightness, contrast, saturation, hue are simulated to retouch the image appearance. Lastly, middle or high level post JPEG compression is enforced. Such augmentations are expected to make the training samples approach the distribution of real-world forgeries. As a result, generalization ability and robustness of the forensic scheme could be enhanced.

*C. Loss Function*

Since malicious tampering typically occurs locally within an image, the tampered region is usually much smaller than the pristine one. Under such an imbalanced distribution, the standard cross-entropy (CE) loss used for tampered/pristine pixel classification concentrates on most negative samples. It may lead to low true positive rate due to misclassification of tampered pixels. In order to attenuate such class imbalance problem, we exploit the focal loss [20] defined as Eq. (4), which can be considered as a general

CE loss.

$$L_{Focal}(y_i, p_i) = -\sum \alpha(1-p_i)^{\gamma} y_i \log(p_i) \\ - \sum (1-\alpha) p_i^{\gamma} (1-y_i) \log(1-p_i) \quad (4)$$

where $y_i$ and $p_i$ are the ground truth and prediction labels for each pixel of the sample image, respectively. The hyperparameters $\alpha$ and $\gamma$ are the weights, for making the model focus on learning from more difficult samples, and control the gradient of different imbalanced samples to a certain extent.

Intersection of Union (IOU) is another popular evaluation metric in tampering localization tasks. To reinforce the focal loss, we apply an optimized IOU measure, i.e., Lovasz-softmax loss [21], as the second loss function term. That is,

$$L_{Lovasz}(y_i, p_i) = \overline{\Delta J}\left(\max(1 - p_i y_i, 0)\right) \quad (5)$$

where $\overline{\Delta J}$ denotes the Lovasz extension to traditional IOU loss [21]. Overall, the final loss function is

$$L(y_i, p_i) = L_{Focal}(y_i, p_i) + \lambda L_{Lovasz}(y_i, p_i) \quad (6)$$

where $\lambda$ is the weight. The focal loss addresses the pixel class imbalance and difficult samples. The Lovasz-softmax loss guides the network trained to locate tampered regions accurately.

## III. EXPERIMENTS

In this section, extensive experiments are performed to test the performance of our proposed tampering localization scheme.

### A. Experimental Setup

**Datasets and Metrics.** Like [8], the training and validation sample images are collected from a public synthetic image dataset [10] and CASIAv2 [22]. The 55848 synthetic images [10] are created by copying and pasting arbitrary COCO object regions [19] into pristine Dresden [23] and MFC18 [24] images. The CASIAv2 dataset owns 5123 realistic forgery images. The total 60971 sample images with tampered pixel location maps are randomly divided into training and validation sets by 9:1. The following data augmentation operations are enforced subsequently to each sample image: resizing with a factor in [0.5, 2], cropping to 512×512 pixels, random horizontal flipping, Gaussian noising and blurring, random adjustment of brightness, contrast, saturation and hue, JPEG compression with Q=71~95. Since sample images have different resolutions, resizing and cropping are always performed. Each of the other operations is applied with a probability of 0.5. To keep consistent with the state-of-the-art works [1, 2-5], five public image tampering datasets including Columbia [25], CASIAv1 [22], DSO [26], NIST [27], and IMD [28] are used as testing sample sets. The F1, IOU and AUC metrics [12] are employed to measure the accuracy of tampering localization by comparing the predicted class with the corresponding ground-truth for each pixel in a test image. Towards real-world evaluation, the fixed threshold of 0.5 is set in calculating F1 and IOU scores.

Table I. Comparison of tampering localization accuracy (%) among different variants of the proposed scheme on CASISv1 dataset.

| Encoder | ResNet | √ | | | | | | |
| --- | --- | --- | --- | --- | --- | --- | --- | --- |
| | ConvNeXt | | √ | √ | √ | √ | √ | √ |
| Decoder | No fusing | | √ | | | | | |
| | $\{X_4, X_3\}$ | | | √ | | | | |
| | $\{X_4, X_3, X_2\}$ | | | | √ | | | |
| | $\{X_4, X_3, X_2, X_1\}$ | √ | | | | √ | √ | √ |
| Loss function | Cross entropy | | | | | √ | | |
| | Combined loss | √ | √ | √ | √ | | √ | √ |
| Augmentation | | | √ | √ | √ | √ | √ | √ |
| F1 | | 31.8 | 54.8 | 53.1 | 54.5 | 53.4 | 51.5 | 58.1 |
| IOU | | 29.1 | 50.7 | 50.0 | 51.5 | 50.0 | 48.7 | 54.8 |
| AUC | | 73.1 | 84.6 | 83.1 | 81.9 | 88.5 | 79.7 | 85.3 |

**Implementation Details.** The proposed tampering localization network is implemented based on MMSegmentation framework [29]. All experiments run on a PC with one NVIDIA TitanXP GPU. Model parameters for the backbone ConvNeXt-B are initialized by the weights pretrained on the ImageNet-22K dataset [30]. AdamW optimizer is adopted. The initial learning rate is set as $1\times10^{-4}$ using linear warmup schedule with 1500 iterations and reduced by a poly policy. The maximum iteration is 160K and the batch size is 4. The related parameters are set empirically as $\alpha$=0.5, $\gamma$=2, $\lambda$=1.

*B. Ablation Studies*

**Performance via encoder-decoder configurations.** As for the encoder, ResNet-101 [31] which has a similar architecture to ConvNeXt is compared. In Table I, the test results on CASIAv1 dataset show that ConvNeXt outperforms ResNet-101 by 26.3% and 25.7% in F1 and IOU, respectively. In the decoder, fusing the features from all ConvNeXt blocks, i.e., $\{X_4, X_3, X_2, X_1\}$, behaves the best. The performance degrades when merely part of or none features are fused. Such results validate effectiveness of the proposed encoder-decoder network.

**Selection of loss function.** The influence of loss function on the tampering localization performance is tested. The results in Table I show that the performance of combined loss is higher than that of CE loss. It verifies the advantage of combined loss for tampering localization.

**Performance via data augmentation.** The scheme without applying our proposed data augmentation is also tested. Its F1 and IOU values are 51.5%, 48.7%, which are lower than those of the scheme with augmentation by 6.6%, 6.1%, respectively.

Table II. Performance comparison of different tampering localization algorithms on different datasets. Metric values are in percentage. "-" indicates not applicable.

| Algorithm | Columbia | | | CASIAv1 | | | NIST | | | DSO | | | IMD | | | Average | | |
|---|---|---|---|---|---|---|---|---|---|---|---|---|---|---|---|---|---|---|
| | F1 | IOU | AUC | F1 | IOU | AUC | F1 | IOU | AUC | F1 | IOU | AUC | F1 | IOU | AUC | F1 | IOU | AUC |
| ManTraNet [4] | 35.6 | 25.8 | 74.7 | 13.0 | 8.6 | 77.6 | 9.2 | 5.4 | 64.5 | 33.2 | 24.3 | 79.0 | 18.3 | 12.4 | 74.0 | 21.9 | 15.3 | 73.9 |
| Noiseprint [2] | 36.4 | 26.2 | 84.0 | - | - | - | 12.2 | 8.1 | 67.4 | 33.9 | 25.3 | 90.2 | 17.9 | 12.0 | 70.5 | 25.1 | 17.9 | 78.0 |
| DFCN [12] | 41.9 | 28.0 | 62.3 | 18.1 | 11.9 | 72.9 | 8.2 | 5.5 | 64.5 | 32.0 | 21.7 | 64.1 | 23.3 | 16.1 | 77.7 | 24.7 | 16.6 | 68.3 |
| MVSS-Net [5] | 68.4 | 59.6 | 83.0 | 45.1 | 39.7 | 84.5 | 29.4 | 24.0 | 79.1 | 27.1 | 18.8 | 73.2 | 26.0 | 20.0 | 81.7 | 39.2 | 32.4 | 80.3 |
| OSN [6] | 70.7 | 60.8 | 86.2 | 50.9 | 46.5 | 87.3 | 33.2 | 25.5 | 78.3 | 43.6 | 30.8 | 85.4 | 49.1 | 39.2 | 89.7 | 49.5 | 40.6 | 85.4 |
| Proposed | 88.5 | 85.7 | 93.6 | 58.1 | 54.8 | 85.3 | 37.0 | 31.8 | 74.3 | 27.0 | 22.8 | 80.1 | 50.0 | 43.2 | 84.8 | 52.1 | 47.7 | 83.6 |

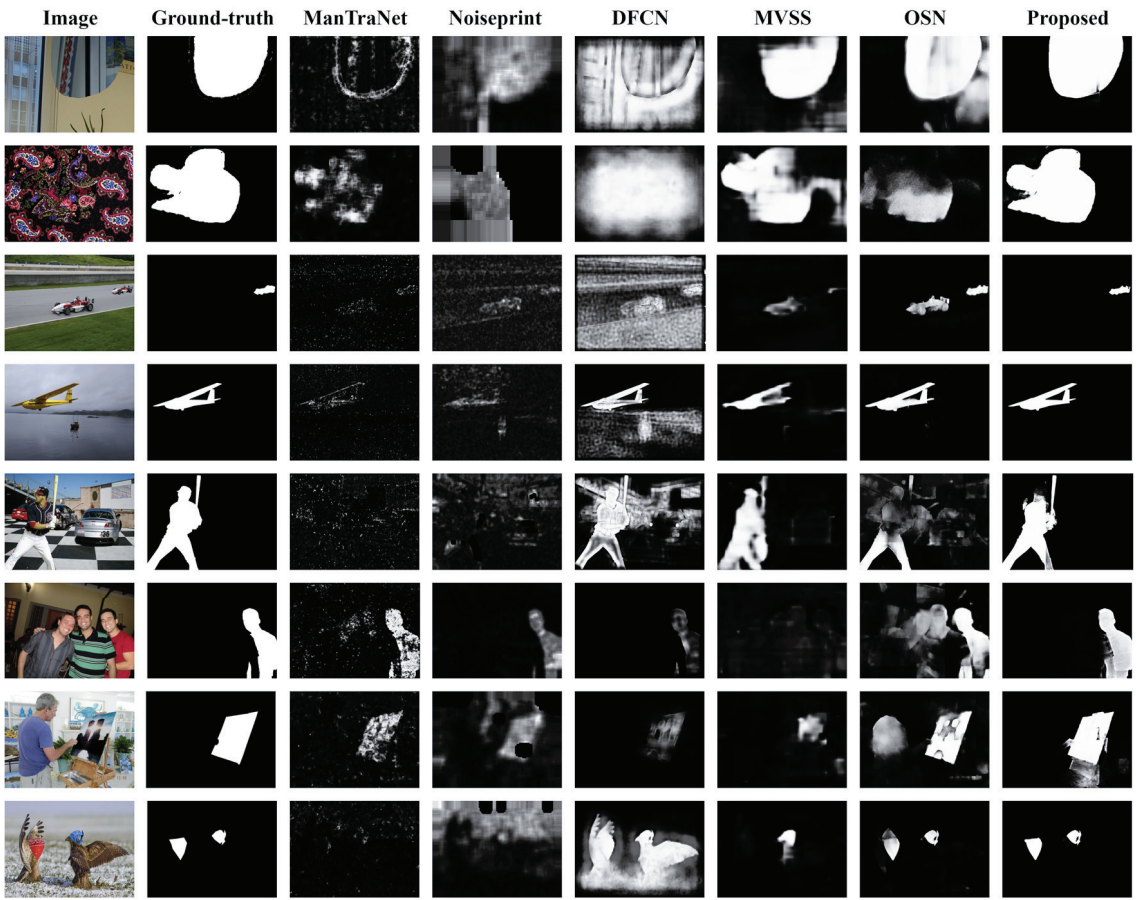

Fig. 2. Tampering localization maps on eight example forgery images. From left to right: test images, ground-truth, and results of different localization methods.

## C. Comparison with Other Localization Algorithms

Performance of the proposed tampering localization scheme is compared with other state-of-the-art ones, i.e., ManTraNet [4], Noiseprint [2], DFCN [12], MVSS-Net [5], OSN [6]. For a fair comparison, we use DFCN model re-trained on our training dataset. The results of other compared methods are either cited from their original papers or gained by running officially released codes. Note that ManTraNet could not infer some large NIST and IMD images with full resolution due to GPU memory constraints. Following the prior work [8], such images are cropped to 2048×1440 pixels for testing. The results in Table II indicate that our proposed scheme achieves the highest localization accuracy on Columbia, CASIAv1, NIST, and IMD datasets. On the average, F1 and IOU achieved by the proposed scheme are respectively 52.1% and 47.7%, which are higher than the second-best ones, i.e., OSN, by 2.6% and 7.1%. Such two algorithms behave rather better than the other ones. It should be mentioned that our scheme meets challenge on DSO dataset, which consists of close-up human faces or body-spliced images. The unromantic performance might attribute to the mismatching between distributions of training and testing samples, which is also a challenge in many pattern classification tasks.

As qualitative evaluation, tampering localization results on eight example forgery images are illustrated in Fig. 2. We can see that most of tampered regions could be identified accurately by our method, which outperforms the other counterparts evidently in terms of completeness and false alarm errors of the identified tampered regions.

Table III. Robustness comparison of different tampering localization algorithms against postprocessing of different online social networks (OSNs). Metric values are in percentage. "-" indicates not applicable.

| Algorithm | OSNs | Columbia | | | CASIAv1 | | | NIST | | | DSO | | | Average | | |
|---|---|---|---|---|---|---|---|---|---|---|---|---|---|---|---|---|
| | | F1 | IOU | AUC | F1 | IOU | AUC | F1 | IOU | AUC | F1 | IOU | AUC | F1 | IOU | AUC |
| ManTraNet [4] | Facebook | 10.1 | 5.5 | 62.4 | 10.1 | 6.4 | 76.4 | 9.2 | 5.5 | 65.4 | 10.8 | 7.1 | 64.5 | 10.0 | 6.1 | 67.2 |
| Noiseprint [2] | | 19.8 | 12.2 | 70.0 | - | - | - | 6.1 | 3.7 | 56.8 | 13.0 | 8.2 | 77.8 | 13.0 | 8.0 | 68.2 |
| DFCN [12] | | 31.5 | 21.4 | 69.6 | 16.1 | 10.2 | 71.7 | 11.6 | 7.7 | 70.3 | 4.9 | 3.0 | 44.6 | 16.0 | 10.6 | 64.0 |
| MVSS-Net [5] | | 69.1 | 60.3 | 83.4 | 38.7 | 33.4 | 83.0 | 26.4 | 21.3 | 78.3 | 27.7 | 19.3 | 73.7 | 40.5 | 33.6 | 79.6 |
| OSN [6] | | 71.4 | 61.1 | 88.3 | 46.2 | 41.7 | 86.2 | 32.9 | 25.3 | 78.3 | 44.7 | 32.0 | 85.9 | 48.8 | 40.0 | 84.7 |
| Proposed | | 87.3 | 84.1 | 92.9 | 53.4 | 49.6 | 83.5 | 36.2 | 31.1 | 73.8 | 26.0 | 21.6 | 80.1 | 50.7 | 46.6 | 82.6 |
| ManTraNet [4] | Weibo | 10.3 | 5.6 | 62.0 | 9.8 | 6.3 | 75.5 | 8.7 | 5.2 | 67.4 | 5.7 | 3.6 | 60.6 | 8.7 | 5.2 | 66.4 |
| Noiseprint [2] | | 16.9 | 10.4 | 63.7 | - | - | - | 5.3 | 3.2 | 57.2 | 7.3 | 4.3 | 62.3 | 9.9 | 6.0 | 61.0 |
| DFCN [12] | | 17.2 | 10.7 | 69.5 | 15.9 | 10.1 | 72.3 | 7.5 | 5.0 | 69.7 | 5.6 | 3.2 | 51.6 | 11.5 | 7.2 | 65.8 |
| MVSS-Net [5] | | 68.9 | 60.1 | 83.2 | 40.3 | 35.3 | 82.4 | 25.1 | 20.0 | 78.3 | 25.8 | 18.3 | 72.3 | 40.0 | 33.4 | 79.0 |
| OSN [6] | | 72.4 | 62.6 | 88.3 | 46.6 | 42.1 | 85.8 | 29.4 | 21.9 | 78.0 | 37.0 | 25.3 | 80.8 | 46.3 | 38.0 | 83.2 |
| Proposed | | 88.2 | 85.3 | 93.4 | 52.9 | 49.7 | 83.1 | 35.7 | 30.8 | 73.5 | 26.2 | 21.6 | 79.7 | 50.7 | 46.8 | 82.4 |
| ManTraNet [4] | Wechat | 19.9 | 12.5 | 61.5 | 8.0 | 4.8 | 73.3 | 9.4 | 5.5 | 65.8 | 6.3 | 3.8 | 58.5 | 10.9 | 6.6 | 64.8 |
| Noiseprint [2] | | 18.8 | 11.2 | 64.4 | - | - | - | 3.8 | 2.2 | 56.7 | 9.2 | 5.6 | 62.1 | 10.6 | 6.4 | 61.0 |
| DFCN [12] | | 40.4 | 27.8 | 68.9 | 19.6 | 12.6 | 72.6 | 5.0 | 3.2 | 70.1 | 16.7 | 10.4 | 55.8 | 20.4 | 13.5 | 66.8 |
| MVSS-Net [5] | | 69.0 | 60.3 | 83.3 | 24.8 | 20.9 | 75.5 | 21.2 | 16.5 | 77.8 | 21.4 | 15.0 | 72.2 | 34.1 | 28.2 | 77.2 |
| OSN [6] | | 72.7 | 63.1 | 88.3 | 40.5 | 35.8 | 83.3 | 28.6 | 21.4 | 76.4 | 36.6 | 25.2 | 82.3 | 44.6 | 36.4 | 82.6 |
| Proposed | | 88.2 | 85.4 | 93.3 | 35.8 | 32.4 | 75.3 | 34.9 | 29.8 | 73.8 | 23.0 | 18.7 | 79.5 | 45.5 | 41.6 | 80.5 |

*D. Robustness Analysis*

As mentioned in [6, 7], various lossy operations adopted by online social networks pose a great challenge for robust image forgery forensics. We evaluate the robustness of different localization algorithms against the postprocessing incurred by online social network platforms, such as Facebook, Weibo, and Wechat. Table III verifies that the comparative performance advantage of our scheme could still be kept as that without postprocessing. For instance, the average F1 scores of ours (50.7%, 50.7%, 45.5%) are always above those of the OSN algorithm [6] (48.8%, 46.3%, 44.6%) on the three platforms, respectively. Furthermore, the overall performance reduction of our scheme is approximately the smallest in different cases. That could be illustratively confirmed by the results on DSO dataset, where F1 of ours always keeps above 23% while that of other methods fluctuates visibly. Although AUC values of our scheme are slightly lower than those of OSN, the overall outperformance among the state-of-the-art algorithms is still evident. It should be noted that AUC might not be a reasonable metric for tampering localization task due to its severe nonlinearity and neglect of class imbalance.

## IV. CONCLUSION

In this work, we have proposed a novel image tampering localization scheme, which mines the generic inherence of images for forgery localization. Hierarchical tampering features are extracted by ConvNeXt encoder. Such multi-scale features are fused properly in decoder by a top-down pathway for estimating the tampering probability of each pixel. Extensive evaluations have verified the effectiveness and robustness of our proposed scheme. The performance gains achieved by the applied data augmentation and combined loss have also been validated in detail.